\documentclass[10pt,conference]{IEEEtran}
\usepackage{graphicx}
\usepackage{xcolor}
\usepackage{amsmath}
\usepackage{url}
\usepackage{hyperref}

\title{Do Larger Fully Connected Neural Networks Generalize Better?}

\author{
\IEEEauthorblockN{Mohammad Jalili Torkamani}
\IEEEauthorblockA{
\textit{School of Computing} \\
\textit{University of Nebraska--Lincoln} \\
Lincoln, NE, USA \\
mjalilitorkamani2@huskers.unl.edu
}
\and
\IEEEauthorblockN{Suhib Hamid}
\IEEEauthorblockA{
\textit{Agronomy and Horticulture} \\
\textit{University of Nebraska--Lincoln} \\
Lincoln, NE, USA \\
shamid2@huskers.unl.edu
}
\and
\IEEEauthorblockN{Moaz N. Ishag}
\IEEEauthorblockA{
\textit{Biological System Engineering} \\
\textit{University of Nebraska--Lincoln} \\
Lincoln, NE, USA \\
mishag2@huskers.unl.edu
}
}
\date{June 2025}

\begin{document}

\maketitle

\begin{abstract}
Greater neural-network capacity does not necessarily yield better performance on unseen data. This study investigates whether a deeper, fully connected neural-network architecture generalizes better than a shallower architecture on the Fashion-MNIST dataset. We compare two architectures under two learning rates and two L2-regularization settings, producing eight configurations. The architectures contain 95,410 and 561,470 trainable parameters, respectively, an approximately 5.9-fold difference. Checkpoints and the final configuration are selected by validation loss, and all configurations are evaluated on a held-out test set. The selected larger configuration achieves 89.82\% validation accuracy and 88.77\% test accuracy. Selecting the lowest-validation-loss configuration within each architecture yields test accuracies of 88.77\% and 88.52\%, respectively, a difference of 0.25 percentage points. Across matched training settings, the larger architecture achieves higher test accuracy in only one of four comparisons. Class-level errors are concentrated among visually similar garment categories, and qualitative pixel-removal analyses suggest sensitivity to shape and boundaries. Overall, the findings suggest that increasing architectural capacity alone does not guarantee better generalization, highlighting the importance of considering optimization and regularization when selecting neural-network architectures.
\end{abstract}


\section{Introduction}

Selecting an appropriate neural-network architecture requires determining whether increased model capacity improves performance on unseen data. Although greater depth and width enable networks to represent more complex functions, their effects on generalization depend on the training conditions and the patterns learned from data. Consequently, an empirical comparison of network architectures should examine whether increasing architectural complexity improves held-out performance and whether any observed advantage persists across different optimization and regularization settings.

Model performance in artificial intelligence and machine learning is influenced by architectural design, data representation, training procedures, and evaluation strategies. Neural-network and linear-regression approaches have been compared for sewer-pipe condition prediction \cite{mohammadagha2025pipes}, while hyperparameter optimization and cross-validation have been investigated for tree-based classifiers \cite{mohammadagha2025tuning}. Ensafi et al. \cite{ensafi2026development} combined feature selection, class-imbalance handling, and classifier stacking for outpatient appointment prediction. Deep learning approaches have also incorporated temporal-spatial processing and attention mechanisms for EEG classification \cite{amini2025adhdeepnet}, as well as noise injection and multi-task learning for produce-image classification \cite{fard2025noisyvit}. These studies illustrate that predictive performance depends on multiple aspects of model design and training, motivating comparisons that consider architectural capacity alongside optimization and regularization.

Input representation, task formulation, and deployment constraints introduce additional considerations. Mehrnia et al. \cite{mehrnia2024assessing} compared MedSAM1 and MedSAM2 for cardiac MRI segmentation, reporting comparable segmentation accuracy despite differences in prompting requirements and processing time. Other studies have incorporated process parameters into transformer-based prediction of ferroelectric polarization \cite{arzefouni2026hzo} and employed Transformer-based reinforcement learning for scheduling larger problem instances without retraining \cite{ardali2026deep}. Deployment constraints have been examined through split computation between wearable devices and remote servers \cite{sazonov2026jitai}, while generative approaches have been used to augment small, imbalanced image datasets \cite{khazrak2025addressing}. Beyond conventional accuracy, open-set intrusion detection research has examined the trade-off between identifying unseen attacks and maintaining precision \cite{safari2025enhancing}, whereas research on generative AI adoption for tourism planning has investigated perceived usefulness, attitudes, and misinformation concerns \cite{safari2026risk}. Together, these studies demonstrate that the evaluation of AI systems depends on their design choices and intended applications.

Evaluation criteria also extend beyond predictive accuracy. Language-model research has examined communication under preference misalignment \cite{balyani2026truthful} and identity-conditioned interaction quality \cite{amirimargavi2026fairness}. Similarly, research on AI-supported decision-making emphasizes appropriate reliance on model recommendations \cite{info17101005}. While these studies address different aspects of AI performance, the present work focuses specifically on generalization measured through held-out classification accuracy.

The objective of this study is to investigate whether a deeper, fully connected neural-network architecture generalizes better than a shallower architecture on Fashion-MNIST, a ten-class benchmark of grayscale clothing images \cite{xiao2017fashionmnistnovelimagedataset}. We compare a Basic architecture with one hidden layer against an Extended architecture with three hidden layers. The architectures are evaluated under matched learning-rate and L2-regularization settings using Adam optimization \cite{kingma2014adam}. Validation loss guides model selection, while a separate test set is reserved for evaluating generalization. Since the architectures differ in both depth and width, the comparison examines their overall performance rather than isolating the effect of network depth alone.

\section{Problem Definition}

We consider supervised ten-class classification using \(28\times28\) grayscale inputs with pixel intensities in \([0,255]\). Each input is normalized to \([0,1]\) and flattened into a 784-dimensional vector before being provided to the network. The network produces a probability distribution over the ten target classes using a Softmax output layer, and the class with the highest predicted probability is selected as the output. Model parameters are learned by minimizing categorical cross-entropy between the predicted probability distribution and the one-hot encoded ground-truth label.
\section{Experimental Design}

\subsection{Data and Preprocessing}

We use Fashion-MNIST \cite{xiao2017fashionmnistnovelimagedataset}, which contains 60,000 training examples and 10,000 test examples. The original training set is partitioned into 54,000 training examples and 6,000 validation examples using the fixed TensorFlow Datasets slices \texttt{train[:90\%]} and \texttt{train[90\%:]}, respectively. This is a partition of the dataset's provided ordering rather than a separately randomized or class-stratified split. The separate 10,000-example test set is retained exclusively for final evaluation and does not influence model selection.

Pixel intensities are normalized from \([0,255]\) to \([0,1]\), and each \(28\times28\) input is flattened into a 784-dimensional vector before being provided to the fully connected network.

\subsection{Architectures and Training}
We compare two fully connected neural-network architectures that differ in depth, width, and total parameter count. Table~\ref{tab:architectures} summarizes their structural characteristics. Both architectures receive a 784-dimensional normalized input vector, use ReLU activation functions in their hidden layers, and produce class probabilities through a ten-unit Softmax output layer.

The Basic architecture contains one hidden layer with 120 units, resulting in 95,410 trainable parameters. The Extended architecture contains three hidden layers with 500, 300, and 60 units, respectively, resulting in 561,470 trainable parameters. Thus, the Extended architecture has approximately 5.9 times as many trainable parameters as the Basic architecture. This comparison allows us to examine the generalization performance of architectures with substantially different depths and capacities under matched training settings.

L2 regularization is applied only to the weight kernels of the hidden Dense layers; biases and the output-layer kernel are not regularized. The optimized objective is mean categorical cross-entropy plus \(\lambda\sum_{\ell}\lVert W_{\ell}\rVert_F^2\), summed over hidden-layer kernels. All reported training, validation, and test losses exclude this penalty and measure cross-entropy alone.

Training examples are shuffled with a buffer of 1,024 before batching. The supplied implementation does not set an explicit random seed, so initialization and training order are not fixed for exact replication. Each configuration represents one training run.

All models are optimized using Adam \cite{kingma2014adam} with \(\beta_1=0.9\) and \(\beta_2=0.999\). We use a batch size of 64 and train each configuration for at most 40 epochs.

Each architecture is evaluated using learning rates \(\eta\in\{0.001,0.0005\}\) and L2-regularization coefficients \(\lambda\in\{0,0.0001\}\). Combining two architectures, two learning rates, and two regularization settings produces eight configurations.

Early stopping monitors validation cross-entropy with a patience of three epochs. For each run, the weights corresponding to the minimum validation loss are restored. The final configuration is selected by the lowest validation cross-entropy across all eight runs. For the architecture-level comparison, we additionally identify the configuration with the lowest validation cross-entropy within each architecture. Both selections use validation results alone; test results are used only for final evaluation.

\begin{table}
\centering
\caption{Architectural comparison of the Basic and Extended fully connected neural networks.}
\label{tab:architectures}
\small
\setlength{\tabcolsep}{5pt}
\begin{tabular}{lcc}
\hline
Property & Basic & Extended \\
\hline
Input dimensions & 784 & 784 \\
Hidden layers & 1 & 3 \\
Hidden units & 120 & 500, 300, 60 \\
Hidden activation & ReLU & ReLU \\
Output units & 10 & 10 \\
Output activation & Softmax & Softmax \\
Trainable parameters & 95,410 & 561,470 \\
Relative parameter count & 1.0$\times$ & 5.9$\times$ \\
\hline
\end{tabular}
\end{table}

\subsection{Generalization Evaluation}

We evaluate generalization using the held-out test set, which remains isolated from model training and model selection. Let \(E\) denote the number of incorrectly classified test examples and \(N=10,000\) the total number of test examples. The observed test-error rate is

\[
\hat{e}=\frac{E}{N},
\]

which is equivalent to one minus test accuracy.

We additionally report a 95\% Wilson score confidence interval for the test-error proportion. The Wilson interval avoids relying on the symmetric bounds of the normal approximation and provides an interval within the valid probability range. With \(z=1.96\), the interval is computed as follows:

\begin{equation}
\frac{\hat e+\frac{z^2}{2N}
\;\pm\;
z\sqrt{\frac{\hat e(1-\hat e)}{N}+\frac{z^2}{4N^2}}}
{1+\frac{z^2}{N}}.
\end{equation}

\section{Results}

\begin{table*}[t]
\centering
\caption{Performance of the eight evaluated configurations. Losses are cross-entropy without the L2 penalty, measured at restored checkpoints. Epochs denote total training epochs; error intervals are 95\% Wilson intervals for test error.}
\label{tab:experimental-results}
\small
\setlength{\tabcolsep}{5pt}
\begin{tabular}{lrrrrrrrr}
\hline
Model & Epochs & Regularization & Learning Rate & Val. CE loss & Val. acc. (\%) & Test CE loss & Test acc. (\%) & Error CI (\%) \\
\hline
Basic-1 & 14 & 0.0 & 0.001 & 0.2995 & 89.35 & 0.3409 & 88.22 & [11.16, 12.43] \\
Basic-2 & 20 & 0.0001 & 0.001 & 0.2975 & 89.47 & 0.3323 & 88.06 & [11.32, 12.59] \\
Basic-3 & 15 & 0.0 & 0.0005 & 0.3077 & 88.87 & 0.3408 & 87.68 & [11.69, 12.98] \\
Basic-4$^{\dagger}$ & 21 & 0.0001 & 0.0005 & 0.2923 & 89.42 & 0.3300 & 88.52 & [10.87, 12.12] \\
Extended-1 & 8 & 0.0 & 0.001 & 0.3006 & 88.65 & 0.3531 & 87.56 & [11.81, 13.10] \\
Extended-2 & 9 & 0.0001 & 0.001 & 0.2992 & 88.78 & 0.3435 & 87.63 & [11.74, 13.03] \\
Extended-3$^*$ & 13 & 0.0 & 0.0005 & 0.2817 & 89.82 & 0.3300 & 88.77 & [10.62, 11.86] \\
Extended-4 & 9 & 0.0001 & 0.0005 & 0.3000 & 89.02 & 0.3379 & 87.90 & [11.48, 12.75] \\
\hline
\end{tabular}

\vspace{1mm}
\footnotesize
$^*$ Lowest validation loss overall and within Extended (0.2817). $^{\dagger}$ Lowest validation loss within Basic (0.2923). CE denotes cross-entropy. Restored checkpoint epochs for Basic-1--4 are 11, 17, 12, and 18; those for Extended-1--4 are 5, 6, 10, and 6.
\end{table*}

\begin{figure*}[t]
\centering
\includegraphics[width=0.9\textwidth]{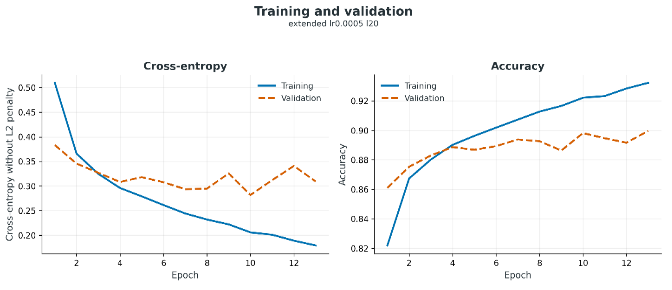}
\caption{Training and validation cross-entropy loss and classification accuracy for the selected Extended-3 configuration.}
\label{fig:training}
\end{figure*}

\subsection{Quantitative Results}

Table~\ref{tab:experimental-results} summarizes the performance of the eight evaluated configurations. Extended-3, the larger architecture with three hidden layers, achieved the lowest validation cross-entropy of 0.2817 using a learning rate of 0.0005 without L2 regularization. It was therefore selected for the final analysis. This configuration obtained 89.82\% validation accuracy and 88.77\% test accuracy, corresponding to a test-error rate of 11.23\% with a 95\% Wilson confidence interval of 10.62--11.86\%.

For the Basic architecture, Basic-4 achieved the lowest validation cross-entropy of 0.2923, whereas Extended-3 achieved the lowest value within the Extended architecture and across all eight configurations. Comparing these validation-selected configurations yields test accuracies of 88.52\% and 88.77\% for Basic and Extended, respectively. Thus, the larger architecture achieved an observed improvement of 0.25 percentage points despite having approximately 5.9 times as many trainable parameters. However, this difference does not establish statistical significance or account for variability across repeated training runs. Furthermore, the Extended architecture achieved higher test accuracy in only one of the four matched learning-rate and regularization settings.

The relative performance of the two architectures also varied with the training configuration. Without L2 regularization, reducing the learning rate from 0.001 to 0.0005 increased the Extended architecture's test accuracy from 87.56\% to 88.77\%, while decreasing the Basic architecture's test accuracy from 88.22\% to 87.68\%. Regularization also produced different outcomes across the architectures. For Extended at the lower learning rate, introducing L2 regularization reduced validation accuracy from 89.82\% to 89.02\%. These observations indicate that the relative performance of the architectures varies with optimization and regularization settings rather than consistently favoring the larger model.

Figure~\ref{fig:training} presents the training and validation performance of the selected Extended-3 configuration. Training loss continues to decrease, whereas validation loss fluctuates after its initial decline. The increasing separation between training and validation performance suggests mild overfitting, as continued optimization does not consistently improve validation performance. Early stopping addresses this behavior by restoring the model parameters associated with the minimum validation loss.

\begin{figure}[t]
\centering
\includegraphics[width=\columnwidth]{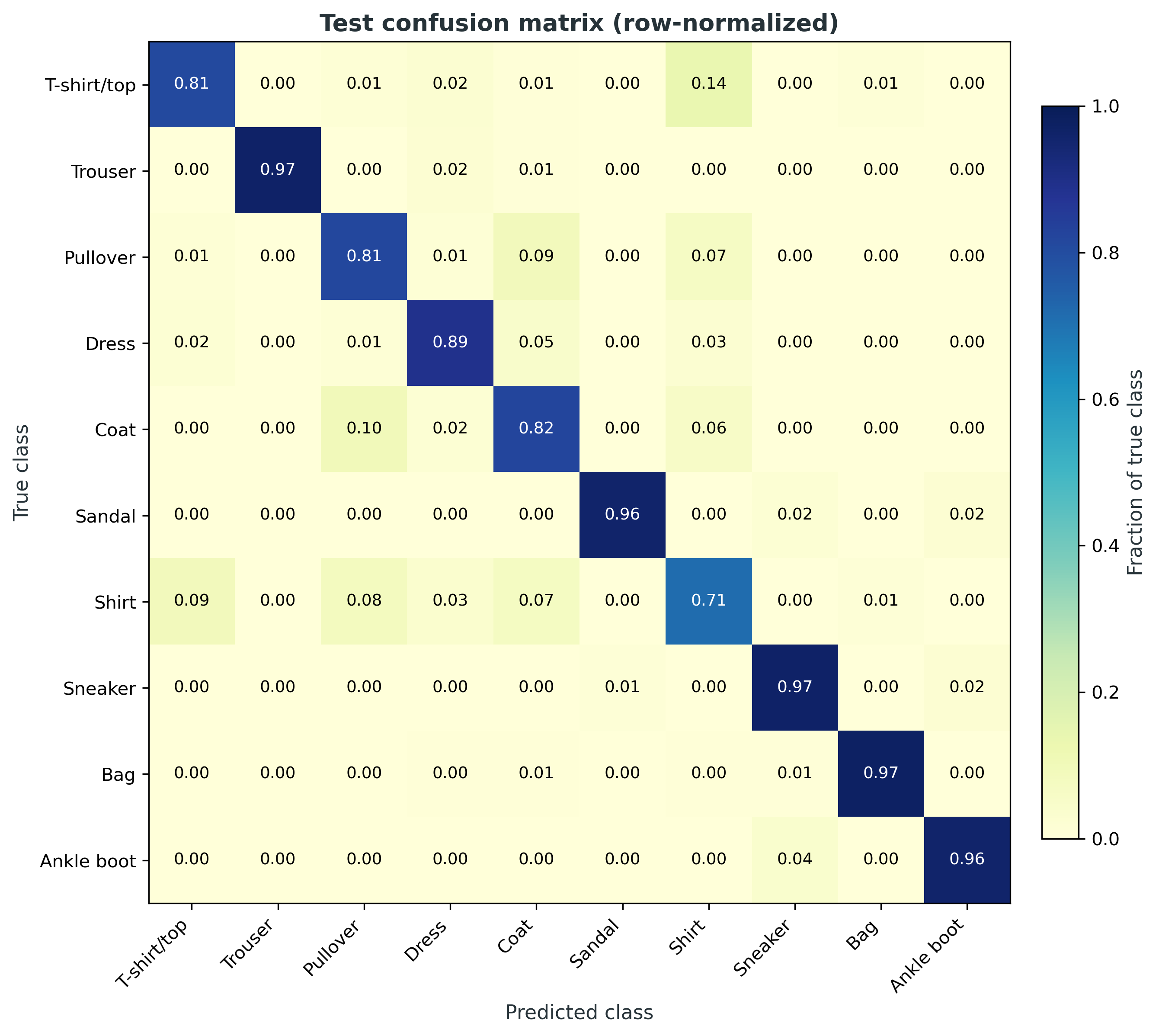}
\caption{Row-normalized test confusion matrix for the selected Extended-3 configuration. Rows represent ground-truth classes and columns represent predicted classes.}
\label{fig:cm}
\end{figure}

The row-normalized confusion matrix in Fig.~\ref{fig:cm} provides further insight into class-level performance. Extended-3 achieves class accuracies of 0.97 for Trouser, Sneaker, and Bag, which have comparatively distinctive visual characteristics. In contrast, Shirt exhibits substantially lower accuracy at 0.71. Misclassified Shirt examples are most frequently assigned to T-shirt/top (0.09), Pullover (0.08), or Coat (0.07). These errors occur primarily among visually similar upper-body garment categories, suggesting that the fully connected architecture encounters greater difficulty distinguishing classes with similar overall shapes and structures.

\subsection{Pixel-Level Influence Analysis}

To further investigate the selected Extended-3 model's predictions, we conduct a leave-one-pixel-out influence analysis on representative test examples. Each pixel is individually set to zero, and the resulting change in the probability assigned to the model's original predicted class is measured. A positive influence value indicates that removing the corresponding pixel decreases the probability assigned to that class.

\begin{figure}[t]
\centering
\includegraphics[width=0.50\textwidth]{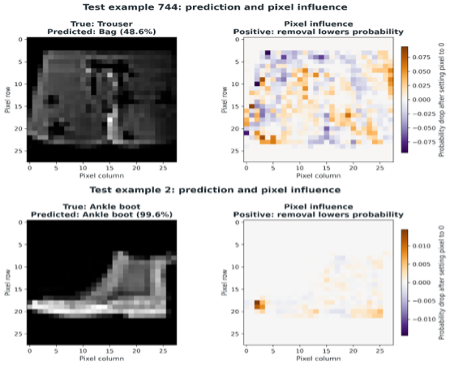}
\caption{Representative test predictions and leave-one-pixel-out influence maps for Extended-3. Top: a Trouser misclassified as a Bag. Bottom: a correctly classified Ankle boot. Positive influence indicates that removing a pixel decreases the probability assigned to the original predicted class.}
\label{fig:saliency}
\end{figure}

Figure~\ref{fig:saliency} illustrates one incorrect and one correct prediction. In the first example, a Trouser is misclassified as a Bag with a predicted probability of 48.6\%. The corresponding influence map highlights pixels across both the outline and interior of the image, consistent with the folded garment's broad, bag-like appearance. In the second example, an Ankle boot is correctly classified with a predicted probability of 99.6\%, with influential pixels concentrated primarily around the toe and heel regions.

These influence maps provide qualitative insights into individual predictions rather than evidence of model-wide behavior. Nevertheless, the examples suggest that shape- and boundary-related information contributes to the selected model's classification decisions.

\section{Discussion and Limitations}

The larger architecture achieved the highest observed test accuracy, but its advantage was not consistent across the matched training settings. Although the validation-selected configurations differed by 0.25 percentage points in test accuracy, this difference does not establish statistical significance or reproducibility across training runs. Moreover, the two architectures responded differently to changes in learning rate and L2 regularization, indicating that their relative performance depends on the training configuration. These findings suggest that architectural size should be evaluated alongside optimization and regularization choices rather than considered independently.

Several limitations constrain the generalizability of these findings. First, the evaluation includes only two fully connected architectures, preventing broader conclusions about the relationship between model size and generalization. Second, the reported Wilson confidence intervals quantify uncertainty in individual test-error proportions but do not establish the statistical significance of the difference between models evaluated on the same test examples. Assessing this difference through a paired comparison would require the models' per-example predictions. Third, each hyperparameter configuration was evaluated in a single training run. Given the stochastic nature of neural-network optimization, repeated experiments across multiple random seeds would provide a more reliable assessment of performance variability. Fourth, the two architectures differ simultaneously in depth, width, and parameter count, making it impossible to isolate the individual contribution of any one of these factors. Future experiments involving approximately capacity-matched architectures could help distinguish the effects of depth from those of width and total parameter count. Finally, both models operate on flattened image inputs without explicitly preserving spatial relationships, limiting the applicability of the findings to architectures specifically designed for structured visual data.

Within the broader AI and machine learning literature, the present study addresses one dimension of model evaluation: predictive performance on a held-out benchmark. Other evaluation frameworks consider aspects beyond classification accuracy. For example, language-model studies have investigated communication under preference misalignment \cite{balyani2026truthful} and demographic differences in interaction quality \cite{amirimargavi2026fairness}. Although these studies address different tasks and evaluation objectives, they illustrate the importance of assessing AI systems through criteria appropriate to their intended applications. In our experiments, the confusion matrix and pixel-removal influence maps complement aggregate accuracy by providing class-level error patterns and qualitative insights into individual predictions. However, evaluating additional dimensions, such as robustness and fairness, would require separate experimental designs.

Beyond predictive performance, practical model selection must also account for computational requirements and deployment constraints. For instance, the wearable intervention framework of Sazonov et al. addresses bandwidth, power, and latency through split computation \cite{sazonov2026jitai}, while the NoisyViT study identifies edge-device latency as a deployment limitation \cite{fard2025noisyvit}. These considerations motivate extending future architecture comparisons to include inference latency, memory consumption, and energy requirements alongside predictive accuracy. Because parameter count alone does not determine deployment efficiency, the present results should be interpreted primarily in terms of predictive generalization. Assessing deployment efficiency and user adoption remains outside the scope of this study and represents a direction for future evaluation.
\section{Conclusion}

We compared two fully connected Fashion-MNIST architectures across four matched learning-rate and L2-regularization settings. Despite containing approximately 5.9 times as many parameters, the larger architecture achieved higher test accuracy in only one of those settings. Selecting the lowest-validation-loss configuration within each architecture yields 88.77\% test accuracy for the larger model and 88.52\% for the smaller model. The observed 0.25 percentage-point difference remains subject to test-sample uncertainty and training variability.

For the architectures and single-run experiments considered here, substantially increasing model size did not produce a consistent generalization benefit. Future work should separate the effects of depth, width, and parameter count, repeat training across random seeds, and extend evaluation to additional datasets and computational-cost measures.

\section{Data Availability}
\label{sec:data_available}
The source code, experimental configurations, and artifacts associated with this work are publicly available at
\href{https://github.com/MJaliliT/DLFCN2026} {https://github.com/MJaliliT/DLFCN2026}.

\bibliographystyle{IEEEtran}
\bibliography{ref}

\end{document}